\documentclass[10pt,twocolumn,letterpaper]{article}

\usepackage{cvpr}
\usepackage{times}
\usepackage{epsfig}
\usepackage{graphicx}
\usepackage{amsmath}
\usepackage{amssymb}
\usepackage{wrapfig}
\usepackage{algorithm}
\usepackage{color}
\usepackage{makecell}
\usepackage{booktabs}
\usepackage{subcaption}

\usepackage{etoolbox,lipsum}
\newcommand{\myparagraph}[1]{\vspace{0.1em}\noindent\textbf{#1}}

\usepackage[pagebackref=true,breaklinks=true,letterpaper=true,colorlinks,bookmarks=false]{hyperref}

\cvprfinalcopy 

\def\cvprPaperID{1289} 

\ifcvprfinal\fi
\begin{document}

\title{A Domain Based Approach to Social Relation Recognition}

\author{Qianru Sun \qquad Bernt Schiele \qquad Mario Fritz \\
Max Planck Institute for Informatics, Saarland Informatics Campus\\
{\tt\small \{qsun, schiele, mfritz\}@mpi-inf.mpg.de}}


\makeatletter
\let\@oldmaketitle\@maketitle
\renewcommand{\@maketitle}{\@oldmaketitle
  \includegraphics[width=\linewidth]{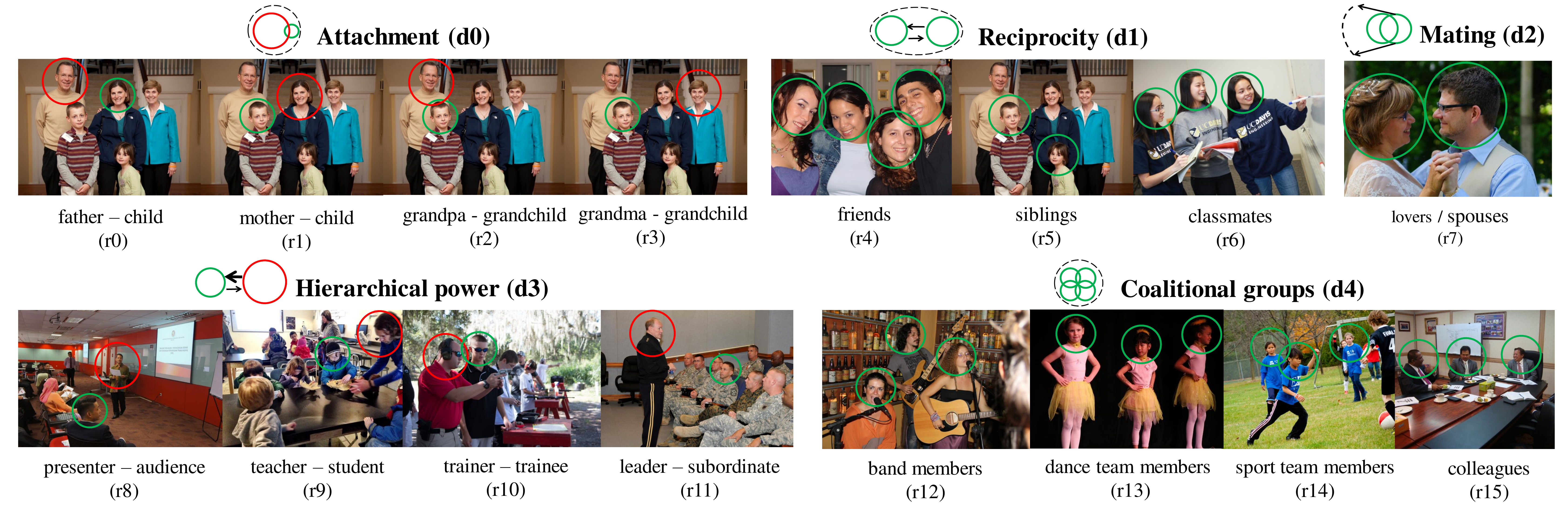}
    \refstepcounter{figure} Figure~\thefigure: We investigate the recognition of social relations in a domain-based approach.  Our study is based on Bugental's social psychology theory \cite{Bugental2000} that partitions social life into 5 domains from which we derive 16 social relations.
    \label{allDomains}
    \bigskip}
\makeatother

\maketitle

\vspace{0.4cm} 
\begin{abstract}
\vspace{-0.4cm} 

Social relations are the foundation of human daily life. Developing techniques to analyze such relations from visual data bears great potential to build machines that better understand us and are capable of interacting with us at a social level.
Previous investigations have remained partial due to the overwhelming diversity and complexity of the topic and consequently have only focused on a handful of social relations. In this paper, we argue that the domain-based theory from social psychology is a great starting point to systematically approach this problem. The theory provides coverage of all aspects of social relations and equally is concrete and predictive about the visual attributes and behaviors defining the relations included in each domain. We provide the first dataset built on this holistic conceptualization of social life that is composed of a hierarchical label space of social domains and social relations. We also contribute the first models to recognize such domains and relations and find superior performance for attribute based features. Beyond the encouraging performance of the attribute based approach, we also find interpretable features that are in accordance with the predictions from social psychology literature. Beyond our findings, we believe that our contributions more tightly interleave visual recognition and social psychology theory that has the potential to complement the theoretical work in the area with empirical and data-driven models of social life.
\end{abstract}

\section{Introduction}

Today, major part of our social life is captured via social media. As we communicate through  multi-modal channels such as Facebook or Twitter, we leave traces that explicitly and implicitly capture social relations in texts, blogs, images and video~\cite{Fairclough2003}.
As we are approaching a future, where intelligent and potential autonomous systems become our assistants and coworkers, we not only want them to be proficient at their task, but also enable them to blend in and act appropriately in different situations of our -- human -- life. Additionally, through better understanding about such hidden information we would like to inform users about potential privacy risks.

Both tasks require conceptualizations and models of social life that provide coverage of all aspects of social life and lend themselves to computational models from sensory input. While initial efforts have been undertaken to address this challenging problem, progress is hampered by the overwhelming diversity and complexity of social life. Most often, computational models to recognize social relations are limited only to a handful of adhoc defined classes.

Therefore, in this paper we start from the social psychology literature to
identify a conceptualization of human social life that is holistic and thereby encapsulates all aspects of interpersonal interaction, but at the same time is concrete and specific, so that a computational model can be build on top of this theoretic underpinning.
We argue to build on Bugental's domain-based theory \cite{Bugental2000} which partitions social life into 5 domains, namely: Attachment, Reciprocity, Mating, Hierarchical power, and Coalitional groups.
Based on these social domains, we derive a list of 16 social relations. This conceptualization of social life is illustrated in Figure \ref{allDomains} with specific photo examples.

In order to study social domains and relations, we build a dataset based on this domain-based theory. We extend the dataset called People in Photo Album (PIPA) \cite{ZhangN2015} by 26,915 person pair annotations. The label space is hierarchical, by assigning social domain labels that partition the data into 5 domain classes as well as assigning a label for the particular relation that two persons appear to be in.

Finally, we build the first computation models based on this theory that predicts social domain and relation from image data. We analyze fully data-driven models as well as  semantic attribute based models whose attributes are based on the domain-based theory.
In experiments, we find that these semantic attributes perform best in our dataset and that in addition lead to interpretability of the empirically found concepts of social life. Importantly, our empirical models correlate with the predictions of the domain-based theory.

Our contribution is three-fold: (1) we adapt the social domain theory as a framework for computer vision based analysis of social relations; (2) by annotating a large scale Flickr photo dataset with both relation and domain labels, we provide a more comprehensive dataset than previous work; (3) we collect semantic attributes from human body and head images according to the social psychology study \cite{Bugental2000}. An in-depth analysis of attribute significance is performed to bridge the gap between social psychology theories and our computational models.

\vspace{-0.1cm}
\section{Related works}
\vspace{-0.1cm}

Social relation is a significant part of social network study \cite{Fairclough2003,Barr2014,LiLJ2015}. This section focuses on the related work in computer vision while the next section outlines different theories in the psychology literature.

\noindent\textbf{Kinship recognition.} Relationships among family members are the most basic social relations for human. There exist a large number of studies about family member recognition and kinship verification \cite{Wang2010,Singla2008,Dai2015,Shao2014,XiaS2012,Guo2014,ChenY2012,Dehghan2014,ZhangZ2015}.
Most of these works focus on familial relations: husband-wife, parents-children, siblings, grandparents-grandchildren. Researchers leverage certain visual patterns exhibited in these relations. For instance, for two people in a wife-husband relation, husband's face is usually in a higher position than wife's \cite{Wang2010,Dai2015}.
Not only the location information but also the facial appearance, attributes and landmarks are essential features to verify family members. Dehghan \etal \cite{Dehghan2014} learn the optimal face features to answer the question of ``Do offspring resemble their parents?''. Singla \etal \cite{Singla2008} propose some attribute-related assumptions, \eg, two people of similar age and opposite gender appearing together are spouses.

Based on the social domain definition in \cite{Bugental2000}, familial relations between adults and offspring are in Attachment domain, for which attribute categories such as age, gender and emotion are essential cues. Sibling relation is categorized in Reciprocity domain, which shows more functional and appearance equality than Attachment domain. This is also consistent with the visual pattern of siblings.

\noindent\textbf{Social role recognition in events.} In social events, there are immediate social roles and inherent relations among participants. The notion of ``social roles'' here models the expected behaviors of certain people \cite{Ramanathan2013,LanT2012,ShuT2015,ZhangJ2011}. For example, in a child birthday party, social roles are birthday child, parents, friends and guests \cite{Ramanathan2013}. Instead of immediate roles, we focus on the identity-specified interpersonal relations, which naturally derive permanent social roles. For example, if ``leader and subordinate'' is confirmed, then it is easy to define the leader's social role as a manager/boss which is much more permanent than ``the guest in a party''. More importantly, our social relation definition is based on psychology studies that suggest comprehensive social scopes in people's long life.

\noindent\textbf{Social categorization and occupation recognition.} Social life endows various social appearances to people. Some research focus on urban tribes in daily life \cite{Murillo2012}, social categories defined by Wikipedia \cite{Kwak2013,ShuH2014}, and popular groups such as ``Loli'', ``Syota'' and ``Goddess'' which are mostly derived from social networks \cite{Hong2015}. These fine-grained categorization uses body/face positions and attributes such as age, face appearance, hair style, clothing style and so on. Occupation recognition studies \cite{Song2011,ShaoM2013} not only use personal attributes but also leverage the contextual information in a semantic level, \eg, a waiter is more likely to stand beside sitting consumers in a restaurant.

\noindent\textbf{Relation prediction.} Social relation is different with the visual relation of ``subject-predicate-object'' \cite{Hangwang2017, Cewu2016}, but it is derived from human social behaviors. A relevant topic is intimacy prediction based on interactive poses \cite{ChuX2015,Yang2012}. In \cite{ChuX2015}, human poses combined with relative distance, leaning direction and orientation are proposed rich representation. Another interesting work is relation traits estimation by faces \cite{ZhangZP2015}. It predicts relation traits such as ``warm'', ``friendly'' and ``dominant'' in face images. Our work is different that we aim to do relation categorization and analyze social domains covering people's social life. The social psychology basis is introduced in the following section.

\section{Social psychology theories}

People organize their social life in terms of their relations with other people \cite{Fiske1992}.
The traditional view is that socialization consists of the individuals' learning of principles that can be applied to all social situations \cite{Reis2000}. Due to the diversity of social situations, it is almost impossible to define a completed list of social relations. We study social psychological theories to identify a theory that: (1) provides a broad coverage of our social life, (2) is concrete enough to allow deriving relevant social relations, and (3) lends itself for computational modeling and recognition in images and video. After reviewing on related theories, we argue that Bugental's social domain theory \cite{Bugental2000} is a suitable candidate.

\subsection{Social domain theory~\cite{Bugental2000}}
\label{socialdomaintheory}

Social domain theory~\cite{Bugental2000} partitions social life into 5 social domains and argues that these cover all relevant aspects of our social interactions.
Additionally, these domains manifest themselves in concrete social behavior that can be recognized from visual data.
Bugental~\cite{Bugental2000} gives comprehensive definitions for each domain including explanations of social cues like appearances and behaviors. Concrete and exemplary social relations are also proposed to illustrate the high-level concept of each domain.
While it is illusive to expect that a comprehensive list of all social relations within a domain can be given, the domain partition along with its clear definition serves as a basis to derive social relations from our dataset (see Section~\ref{labelList}).
Specifically, domain definitions and some of the examples given for social relations are as follows:

\myparagraph{Attachment domain}, characterized by proximity maintenance within a protective relationship, \eg kinship between parents and children. Human attributes such as age difference, proximity and the activity of seeking protection are social cues which can be visually recognizable.

\myparagraph{Reciprocity domain}, characterized by the negotiation of matched benefits with functional equality between people. Key features are the matched and mutually beneficial interactions in long-term accounting process, which are quite common among friends and siblings. Typically, age difference among peers is small, which is an important semantic attribute. Also, mutual activities such as ``gathering'' and ``sharing'' often appear in this domain.
Sequenced exchange of positive effect is another factor, which is hard to predict in an image but might be useful when employing video.

\myparagraph{Mating domain}, concerned with the selection and protection of access to the sexual partner, \eg, the relationship between lovers. Gender cues and the behavior cue of caring offspring are essential for this domain. Bugental also emphasized the facial attractiveness of prospective partners suggesting that facial and most likely also full body appearance are important cues.

\myparagraph{Hierarchical power domain}, characterized by using or expressing social dominance. Dominance appears in resource provision and threatening activities. Concrete examples are leaders, powerful peers and teachers. On the other hand, submissive activities like ``listening'' and ``agreeing'' are more adaptive for those who lack dominance.

\myparagraph{Coalitional groups domain}, concerned with the identification of the lines dividing ``us'' and ``them''. The focus is on the grouping and conformity cues which ranges from colleagues at work, over sport team members to band members. Coalitional group members often share similar or identical clothing and perform joint activities.

\subsection{Related theories}
For completeness, we briefly discuss several related social psychology theories ordered by the time of appearance.

(1) Parson's theory of role expectations \cite{Parson1965} uses five pattern variables to compose a systematic classification of social relations, namely General-categorical, Personal-categorical, General-behavioral and Personal-behavioral.
(2) MacCrimmon and Messick's theory of social motives \cite{MacCrimmon1976} studies individualism and proposes  six motives: 
Altruism, 
Cooperation, 
Individualism, 
Competition 
and Aggression. 
%
(3) Mills and Clark's theory of Communal and Exchange relations \cite{Clark1979},
focuses on the rules and expectations governing ``give'' and ``take'' benefits.
%
(4) Foa and Foa's theory of resource exchange \cite{Foa1980}, defines
social relations based on six social resources: Love, Status, Money, Goods, Services and Information.
%
(5) Fiske's theory of relational models \cite{Fiske1992}
argues that relations can be differentiated into four parts: Communal sharing, Authority ranking, Equality matching, and Market pricing.

Theories (1)-(3) are rather abstract and theoretical in the field of social psychology and thus not concrete enough for our purpose in computer vision. Theory (4) considers social resources which are concrete but are difficult to infer from visual data.
Theory (5) is similar to Bugental's theory but focuses on the cognitive individual experiences reflecting personal history, while Bugental concerned more to link the theory to social behaviors, appearances and environments, which are often visually interpretable. Moreover, Bugental's domain partitioning is based on a large number of social cues (\eg the Table 1 of \cite{Bugental2000}), which helps to devise computational models for his theory.

\begin{figure*}[htp]
    \centering
  \includegraphics[width=\linewidth]{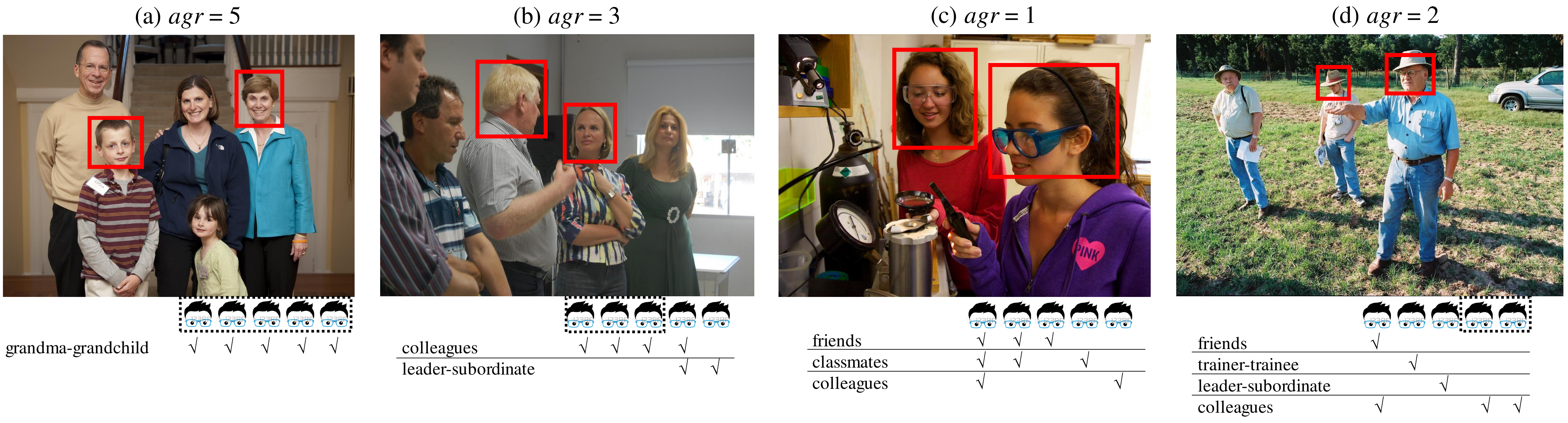}
  \caption{Photo samples of different agreements ($\text{\it agr}$). Annotators in the dashed box are in agreement. (a) is obvious a family photo and contains grandma-grandchild. In (b), the forth annotator gives a vote to colleagues but is not counted in agreement, since agreement means annotators' recognitions are exactly the same. (c)(d) contain ambiguous social relations, thus result in very low agreement.}\label{annotationSamples}
\end{figure*}

\section{Social domain and relation data set}
\label{sec4}

To study social domains and relations, we start from the PIPA dataset~\cite{ZhangN2015}.
The following discusses the dataset,
how we have derived the social relations for annotation
and presents the annotation process itself.

\myparagraph{PIPA dataset.}
The PIPA dataset was collected from Flickr photo albums for the task of person recognition~\cite{ZhangN2015}.
Photos from Flickr cover a wide range of social situations and are thus a good starting point for our study. PIPA contains 37,107 photos with 63,188 instances of 2,356 identities.
For each annotated person, the head bounding box and the identity ID are available.
The same person often appears in different social scenarios and interacting with different
people which make it ideal for our purpose. Identity information is used for selecting person pairs and defining train-validation-test splits (see Section~\ref{sec:dataSplits}).

\subsection{Social relation labels}
\label{labelList}

As argued before, the social domain theory is a good basis to derive social relations for annotation. While the domains are precisely described, we still have to obtain a set of relation labels. We proceed in three steps. \\
{\bf (1)} In~\cite{Bugental2000}. 12 exemplary social relations are listed for the different domains. We takes these as a starting point.\\
{\bf (2)} As this list is unlikely to cover all social relations in PIPA, we start with a pre-annotation phase of PIPA covering 10\% of the data. During this pre-annotation phase, we identify social relations not yet covered and use FrameNet~\cite{framenet} to name the relation and the detailed social domain description to associate the social relation to its domain. This increases the number of relation labels from 12 to 20.\\
{\bf (3)} Finally, We ask five annotators to annotate all of PIPA using the derived list of 20 relations. After the annotation process was concluded, we analyze the statistics and found 4 relations with insufficient occurrences ($\le 20$) or insufficient agreement. This results in our final list of 16 relations. For example, kinship like ``mother-child'' is in Attachment domain, and the relation ``leader-subordinate'' is in Hierarchical power domain.

\subsection{Annotation process}

\myparagraph{Annotators.} Annotating social relations might be subjective and ambiguous. One reason is that a person pair may have multiple plausible relations, as shown in Figure~\ref{annotationSamples}.
Another reason is that the definition of the same social relation might differ,
depending on the cultural backgrounds of the annotators.
We selected five annotators from Asia, Africa, Europe and America and gave them
detailed explanations and photo examples to help them keep some basic consistency (given in supplementary files).

\myparagraph{Annotation Protocol.} Annotators are asked to individually annotate all person pairs for which we present pairs of head bounding boxes.
For each pair the annotator can either pick a relation from our list or, if they are too uncertain, can skip this pair.
For example, two people wearing uniforms and working in the factory should be labeled as ``colleagues'', as the cues of action ``working'', clothing ``uniforms'' and environment ``factory'' are obvious. If the annotators are uncertain they are asked to indicate this by clicking ``maybe'' for this relation. Based on our pre-annotation phase we allowed at most 3 relation labels per person pair which is justified also by the following statistics: 92.3\% of the person pairs have 1 relation annotated, 7.5\% have 2 relations and only 0.3\% have 3 relations.


\subsection{Label statistics}
\label{sec:labelStatistic}

After the annotation process we have 26,915 person pairs annotated by five annotators. Given the fact that each annotator can give up to 3 labels per pair or skip a pair, we obtain a total number of 134,556 annotations. For about 8\% of annotations, the annotators choose ``maybe''. Given that this is a rather small part of the data we include those in the following statistics.

As mentioned before, annotation of social relations using single images might be ambiguous and subjective.
Interestingly, for 53\% of person pairs at most one relation was chosen across annotators. For 38.8\% of pairs two relations, for 7.4\% three relations
and only for 0.8\% four relations are chosen {\it across} annotators.

Three examples where multiple annotations are chosen are shown in Figure~\ref{annotationSamples} (b)(c)(d). For the image in (d), there are four annotations which are all plausible: the two men might be friends, colleagues, in a leader-subordinate relation, or a trainer-trainee relation.
Such highly ambiguous cases however are less prominent in our dataset than one might expect, and  for a significant number of person pairs there are at most two relations chosen which indicates that a visual recognition approach is indeed feasible.

\subsection{Consistency analysis}
\label{sec:consistencyAnalysis}

We define $\text{\it consistency}\in [1, 5]$ to be the level of complete agreement ($\text{\it agr}$) among the 5 annotators. For instance, $\text{\it consistency=} 3$ means $\text{\it agr}\geq 3$ that at least 3 annotators give the exact same labels to a person pair. For examples see figure~\ref{annotationSamples} from left to right:
as all annotators give just one and the same label, the first image has $\text{\it agr=}5$;
the second has $\text{\it agr=}3$ only, as the fourth annotator not only gives the colleagues relation but also a second relation and is thus not in complete agreement with the first three annotators; the third and the fourth image correspond to  $\text{\it agr=}1$ and $\text{\it agr=}2$.
It is noted that $\text{\it agr=}1$ is the lowest possible value as each annotator is always in complete agreement with herself/himself.

\begin{figure}[t]
    \centering
  \includegraphics[width=\linewidth]{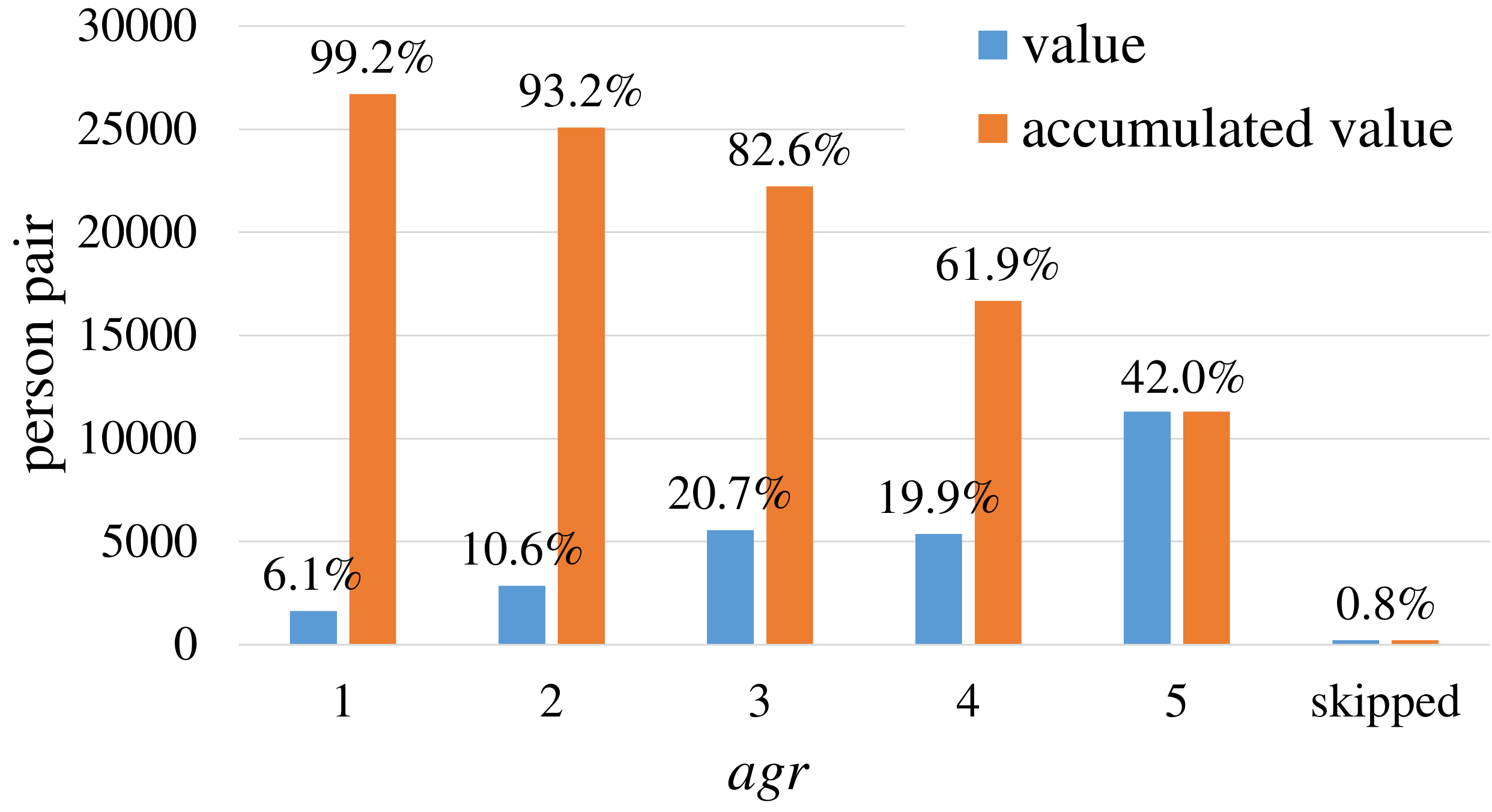}
  \caption{Person pair counting in terms of agreement ($\text{\it agr}$). ``Skipped'' denotes the pairs skipped by all annotators when they couldn't recognize any relation. This paper uses annotations with $\text{\it agr} \geq 3$, i.e., useful annotations take 82.6\% of all person pairs.}
  \label{consistency}
\end{figure}


Figure~\ref{consistency} shows agreement statistics for our dataset.
Even though we require complete agreement as discussed above, for 42\% of our person pairs $\text{\it agr=}5$. This reinforces the observation that the annotations are less ambiguous than one might have expected. In 19.9\% of cases $\text{\it agr=}4$ and in 20.7\% of cases  $\text{\it agr=}3$. Given these encouraging agreement levels we decided to use the annotations, where $\text{\it agr}\geq 3$ as groundtruth. We refer to it as $\text{\it consistency=}3$ in the following, corresponding to $82.6\%$ of our annotations.


\begin{figure*}[htp]
    \centering
  \includegraphics[width=\linewidth]{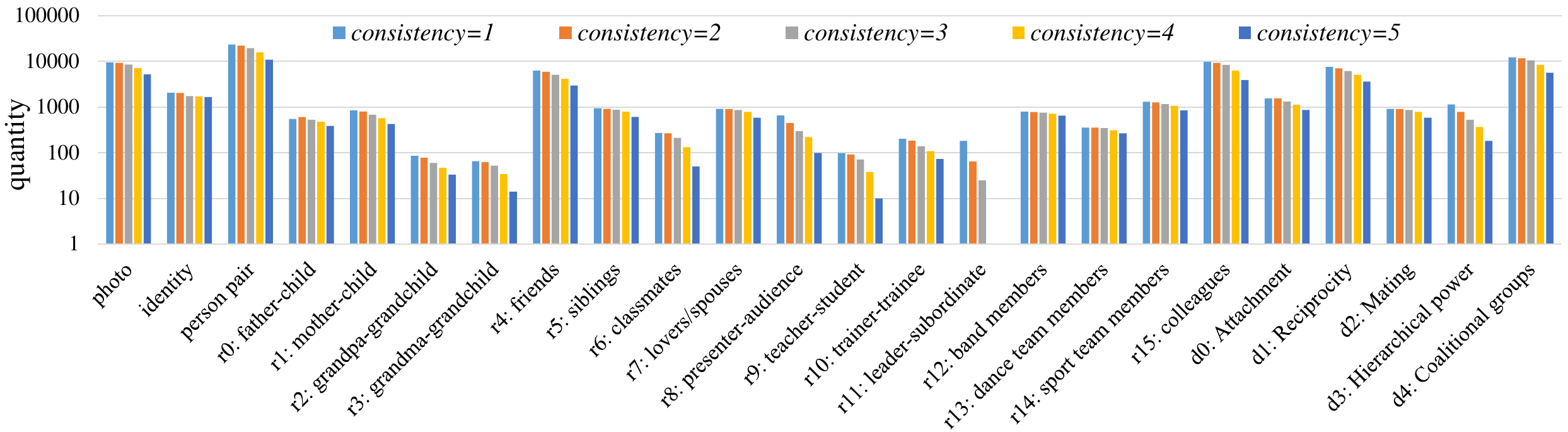}
  \caption{Person pair counting in every social relation (r*) and domain (d*) with $\text{\it consistency}\in [1,5]$. We also list the total counts of photo, identity and person pair. Person pairs with $\text{\it consistency=}3$ (i.e., $\text{\it agr}\geq 3$) are used in our experiments.}
  \label{allNums}
\end{figure*}

In Figure~\ref{allNums}, we show all numbers for photos, identities, person pairs, social relations and domains in terms of $\text{\it consistency}$. As each relation is exactly in one domain, the domain samples is the sum of its relations.

\section{Modeling social domains and relations}

To model semantic head/body attributes, we choose two image regions based on the groundtruth annotation of PIPA: the head region corresponding to the groundtruth annotation and the body region, defined as $3\times$head width, $6\times$head height, following \cite{Oh2015}. For recognition, a pair of head or body regions are fed into the model (see Figure~\ref{doubleStream}). We experiment with two types of models: the first type of models are CNN models trained end-to-end; the second type trains CNN models for semantic attributes derived from the social domain theory, then uses the concatenated feature to learn linear SVM.

\subsection{End-to-end model}

In order to model pairwise relations, we introduce a double-stream CaffeNet, which learns an end-to-end mapping from an image pair to either 5 domain classes or 16 relation classes. This double-stream model is similar to multi-region CNNs \cite{ZhangX2015}, based on LeNet \cite{LeCun1998}. Other similar models can also be considered, such as Siamese-like architectures used for face modeling \cite{Taigman2014,Chopra2005,ZhangZP2015}, and multi-channel CNNs used for person identification \cite{Cheng2016}.

\begin{figure}[h]
    \centering
  \includegraphics[width=\linewidth]{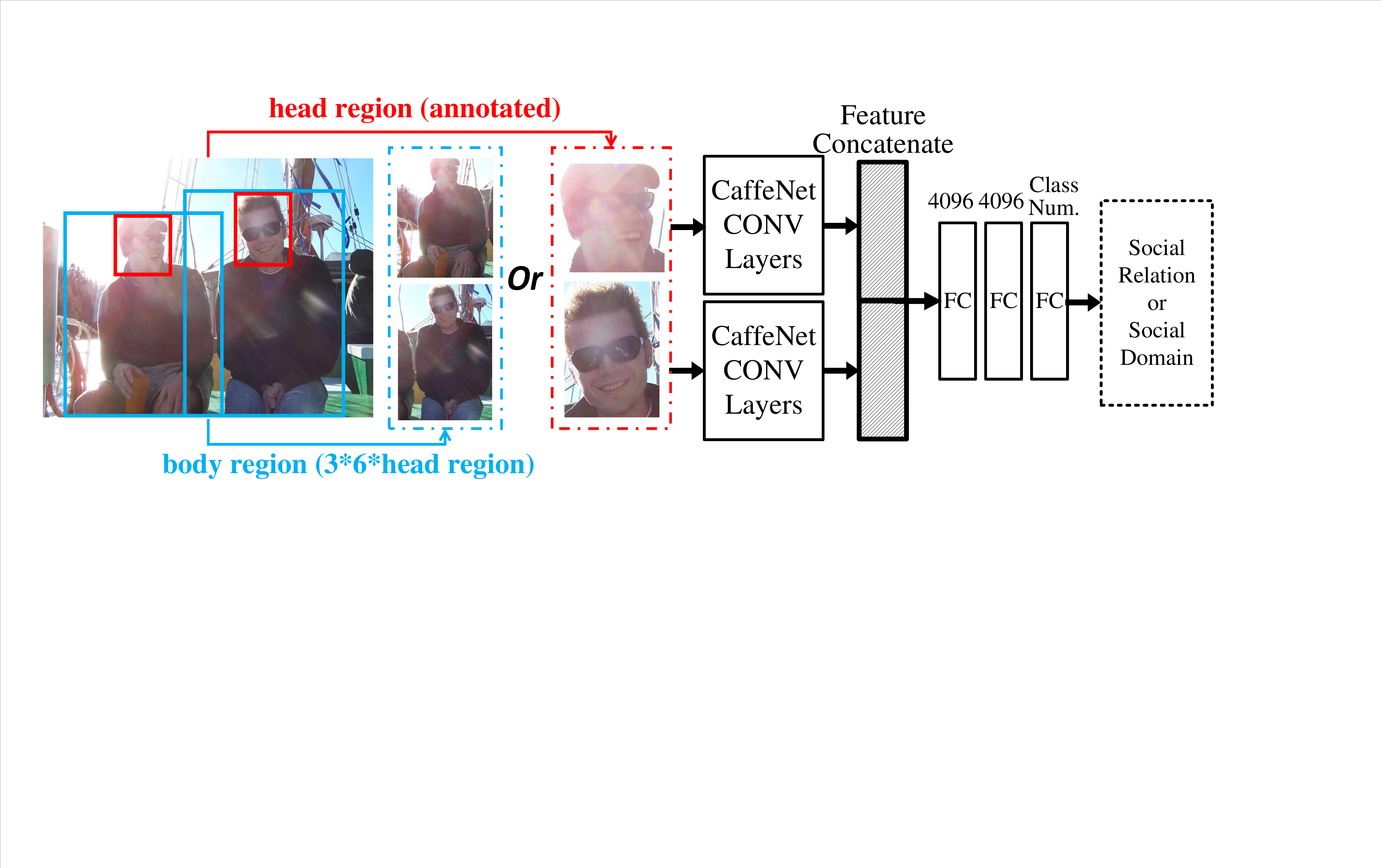}
  \caption{Architecture of double-stream CaffeNet used for modeling social relations and domains. CONV layers are same with those in CaffeNet \cite{Jia2014}. Either head image pair or body image pair are fed into the network. The weights of CONV layers are independently trained for double streams.}\label{doubleStream}
\end{figure}

The double-stream CaffeNet architecture is shown in Figure~\ref{doubleStream}. Each stream has CONV layers as in CaffeNet \cite{Jia2014}. After 5 CONV layers, features are concatenated into one vector that is fed into \emph{fc}6 layer. After \emph{fc}7 layer, we add an \emph{fc}8 to predict either 5 domains or 16 relations.

This model is used to compute baseline recognition results. The same model can be used to predict semantic attributes. This attribute adaption method involves the social cues such as age, emotion and proximity, mentioned in the social psychology article \cite{Bugental2000}.

\subsection{Semantic attributes}
\label{semanticAtts}

The second type of model we propose first predicts an intermediate semantic attribute representation and uses this intermediate representation to predict both social domains and relations. In our view there are three advantages that make this representation attractive.
First, semantic attributes lend themselves for interpretability.
Second, relevant semantic attributes can be derived from the social domain theory and thus can be seen as a way to derive a computational model which in turn allows to link back to the theory.
Third, semantic attributes allow to leverage other datasets annotated with such attributes which we consider crucial as it seems illusive to expect large amounts of training data for recognizing fine-grained social relations.

To start, we collected the semantic head/body attribute categories mentioned in the definition of social domains \cite{Bugental2000}.
For example, in the Attachment domain, a mother and her child have a large age difference and some proximity behaviors, therefore, age and proximity are included. Then, for each attribute category, we either leverage existing datasets to train attribute classifiers using our double-stream CaffeNet (default) or use pre-trained models published in previous works \cite{ChuX2015,imSitu2015}.
Following presents all attributes used.

\myparagraph{Age} infant, child, young, middleAge, senior, unknown are annotated in PIPA by   \cite{Oh2015}. Additionally, we add smallAgeDiff, middleAgeDiff, largeAgeDiff, because age difference is important for distinguishing social relations \cite{Bugental2000}. We use head age and body age respectively trained on the head and body regions of PIPA images.

\myparagraph{Gender} male, female are annotated in \cite{Oh2015}. We add sameGender, diffGender. Same with the age case, we use head gender and body gender.

\myparagraph{Location \& Scale} directly collected from head/body regions of a person pair in PIPA. It is composed of 4-dim location coordinates (x, y, width, height), relativeDistance (far, close) and relativeSizeRatio (large, small).

\myparagraph{Head appearance} 40 classes such as straight hair, wavy hair, wearing earring, wearing hat and so on. This attribute model is trained on the CelebA dataset \cite{celebfaces}, which contains 202,599 head photos of 10,177 identities.

\myparagraph{Head pose, Face emotion}. Poses are frontal, left, right, up and down. Emotions are anger, happiness, sadness, surprise, fear, disgust and neutral. Both models are trained on the IMFDB dataset \cite{IMFDB2015}, which contains 34,512 head images collected from movies.

\myparagraph{Clothing} longHair, glasses, hat, tShirt, shorts, jeans, longPants and longSleeves. We learn the model on Berkeley People Attribute dataset \cite{berkeley2011} containing 8,035 body images, then we extract the features in PIPA using body regions.

Additionally, we utilize the published models of proximity \cite{ChuX2015} and activity \cite{imSitu2015} for extracting body features.

\myparagraph{Proximity} holdingFromBehind, holdingHands, highFive, hug, armOverShoulder, shoulderToShoulder and armInArm. We use the published Multi-task RNN model \cite{ChuX2015}, which was trained on 10,000 images. We first extract the features for PIPA from its \emph{fconv}9 layer which is a 338x50x50 feature tensor containing 338 feature maps. In order to reduce the feature dimension, we use a max pooling operator (to select the most significant feature) along the channel axis, and flatten the feature into 2500 dimensions.

\myparagraph{Activity} 504 activities such as adjusting, ailing, applauding, arranging, attacking, ballooning, baptizing and so on. We use the published CNN-CRF model, which was trained on a dataset of 126,102 images \cite{imSitu2015}. We extract the features for PIPA of the \emph{fc}7 layer which is a 1024-dim feature vector.

\section{Experiments}
\label{sec:experiments}

Our experiments analyze three aspects.
The first set of experiments compare end-to-end training with the proposed semantic attribute model to recognize both social domains and relations.
Additionally, motivated by the fact that it is illusive to have a comprehensive dataset of all relations within a domain, we also analyze domain recognition in a leave-one-relation-out setting that measures domain generalization performance.
Finally, we give insights into our semantic attribute models, in particular in the light of the social domain theory that we base our investigation on.

\subsection{Data splits}
\label{sec:dataSplits}

We introduce two data splitting methods. All-class (AC) splits are used for traditional recognition, where training data cover all classes of social relations (or domains). Single-relation(SR) splits each time leave the data of a relation out of training, then predict the domain label of these data. The goal is to evaluate the model ability to generalize unseen relation classes within a social domain.

\textbf{All-class(AC) splits.} We keep the test split of PIPA dataset unchanged. For train and validation sets, there are strong data imbalance issues on relation classes, as PIPA was originally proposed for person identification \cite{ZhangN2015}.
In order to lessen this problem, we take 3 steps: (1) merge the original train and validation sets; (2) build a new validation set using person pairs from 8 random albums; (3) use the rest data for training. In summary, there are 13,729 person pairs in train set, 709 in validation, and 5,106 in test.

\textbf{Single-relation(SR) splits.} In order to test domain generalization, each time we pick one relation for testing, and randomly divide other data (of 15 relations) into 10 folders by averaging identities. One folder is used for validation, and others for training. Please note, (1) ``10-folder identity averaging'' can not be exactly reached, because an identity may compose multiple person pairs. We discard a pair when one of its identities has samples in train set, meanwhile, the other one has samples in validation set; (2) there is not training data for Mating domain when leaving lovers/spouses out. Therefore, lovers/spouses data are always in train set. Finally, we have 15 groups of train-validation-test splits, thus we run the model for 15 epoches to get testing results.

\subsection{Recognition}

To our knowledge, this is the first work to recognize social domains and test domain generalization across relations. Both are challenging problems. The data of a social domain has large intra-variation, due to diverse relations included. Generalization test is even more challenging, since it is to predict the domain of an unseen relation class. This is difficult but very essential to validate whether social domain theory can cover all its relations, and to what extent our model achieves the coverage of social domain.

To evaluate the semantic attribute based model, we have 5 settings\footnote{The dataset and trained models can be downloaded in our project page: www.mpi-inf.mpg.de/social-relation} as shown in Table~\ref{recog-results}: 
(1) end-to-end double-stream CaffeNet (default) trained from scratch; (2) end-to-end finetuned model from pre-trained in ImageNet; (3) Extract features (from \emph{fc}7 layer) by the ImageNet pre-trained model, then train linear SVM; (4) Replace the pre-trained model in setting (3) with a finetuned model; (5) Concatenate features extracted from attribute finetuned models, then train linear SVM (ours). In setting (5), we test head features, body features and concatenating both.
Except that the initial learning rate $10^{-4}$ is used for scratch, $10^{-5}$ is fixed for others. Each time we run 30 epoches of training samples.

\renewcommand\arraystretch{1.2}
 \begin{table}[h]
 \scriptsize
   \begin{center}
     \label{Atts}
     \begin{tabular}{lccc}
       \toprule
        MODEL &{\sc Relation}& {\sc Domain}  & {\sc Generalization} \\\hline
       {\sc End-to-end scratch}    & 34.4\%        & 41.9\%       & --    \\
       {\sc End-to-end finetuned}  & 46.2\%        & 59.0\%       & 18.5\%\\
       {\sc Pre-trained, SVM}  & 35.9\%        & 53.3\%       & 27.7\%\\
       {\sc Finetuned, SVM}    & 48.6\%        & 63.2\%       & 27.1\%\\ \hline
       {\sc Head attributes, SVM}        & 44.8\%        & 59.4\%       & 21.5\%\\
       {\sc Body attributes, SVM}        & 57.2\%        & 67.7\%       & 32.8\%\\
       {\sc All attributes, SVM}         & 57.2\%        & 67.8\%       & 33.3\%\\
        \bottomrule
     \end{tabular}
       \caption{Accuracies of relation/domain recognition (AC splits), and domain generalization (SR splits). ``ALL'' means concatenating all body and head attribute features.}
       \label{recog-results}
   \end{center}
   \vspace{-0.2cm}
 \end{table}

``End-to-end finetuned'' gets more than 10\% improvement over ``scratch'' for each recognition. However, it fails in the harder task of generalization test (last column), since 18.5\% is around chance level of 20\%. Using semantic attribute based models, we get the highest 14.8\% improvement over ``end-to-end finetuned'' for the generalization test. In the recognition tasks, our best results are 57.2\% and 67.8\%, respectively 8.6\% and 4.6\% higher than best baseline numbers. On the one hand, recognizing relations is much harder due to the larger class number than domains (16 vs. 5). Our attribute model gains a larger improvement for this harder task.
On the other hand, the gap between recognizing relation and domain is not very significant (only 10.6\%), due to the fact that the intra-variance of a domain is quite larger than that of a relation. In particular, Hierarchical power is the hardest domain to recognize. Its relations such as ``teacher-student'' and ``leader-subordinate'' are quite different in both behaviors and appearances.


We can conclude from these improvements that semantic attributes proposed in the social psychology study are very helpful to model high-level social concepts, even though half of the attribute models were trained on other datasets. In the next section, we further analyze our attribute model to get insights into the contribution of specific attributes.

\subsection{Analysis of semantic attributes}
\label{sec:semattributes}

Firstly, we compare the contribution to the overall performance of each attribute category. Then, we present qualitative examples to understand which detail attributes help to improve the recognition.

\begin{figure}[htp]
    \centering
  \includegraphics[width=\linewidth]{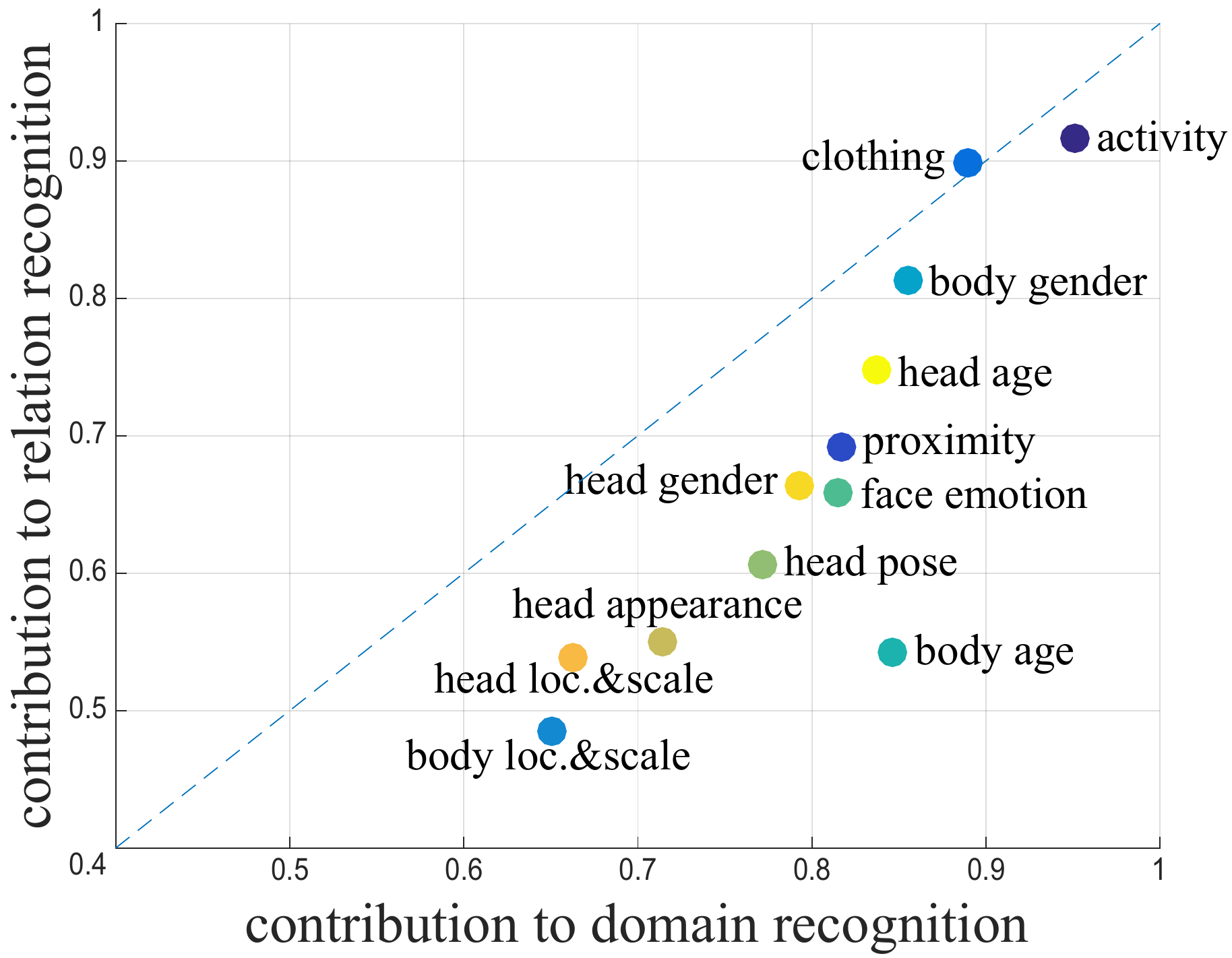}
  \caption{Normalized contributions from 12 attributes.}\label{baseline-C}
  \label{normalizedcor}
\end{figure}

\textbf{Attribute categories}. In Figure~\ref{normalizedcor}, we present the relative recognition contribution of each single attribute category in relation vs. domain models. Taking body age as an example, its X-Y coordinates is computed as follows: (1) we train a model using only one feature: bodyAge; (2) we evaluate the performance for relation and domain, denoted as $\text{\it acc}$(bodyAge, relation) and $\text{\it acc}$(bodyAge, domain). Accuracies of using all attributes, 52.7\% and 67.8\% in Table~\ref{recog-results}, are denoted as $\text{\it acc}$(all, relation) and $\text{\it acc}$(all, domain); (3) normalized results of $\text{\it acc}$(bodyAge, domain)$/\text{\it acc}$(all, domain) and $\text{\it acc}$(bodyAge, relation)/$\text{\it acc}$(all, relation) are used as X, Y coordinates, respectively.

Overall, we can observe in Figure~\ref{normalizedcor} that most attributes are below the diagonal. This indicates that the relative, individual contribution of attributes is stronger for recognizing domains. We conclude that currently more attributes are needed for the relation classification as it is a more challenging task due to more classes and finer granularity.

In terms of the attribute contribution, activity and clothing are ranked in top 2 for both relation and domain (upper right corner). This is consistent to our social domain interpretation in Section~\ref{socialdomaintheory} that social relations regulate our behavior and ``communicate'' relations via appearance. For example, band team members and dance team members in Coalitional groups share similar or identical clothing and perform joint activities both in the level of the specific relation and in the level of the shared domain.
Although these two attribute models are transferred from other datasets \cite{imSitu2015,berkeley2011}, they still make greater influence than age and gender which are learned on PIPA dataset. This reflects that our computational model successfully transfer the social cues from social psychology study to visual data. In terms of the third and forth contributors, age and gender, we can see body gender and head age are relatively important. This is reasonable that in the social psychology definition of Mating domain and Attachment domain, age and age difference, gender and gender difference are dominant features. Another indication is that age is better learned in head images and gender is better learned in body images.

\begin{figure}[t]
    \centering
  \includegraphics[width=\linewidth]{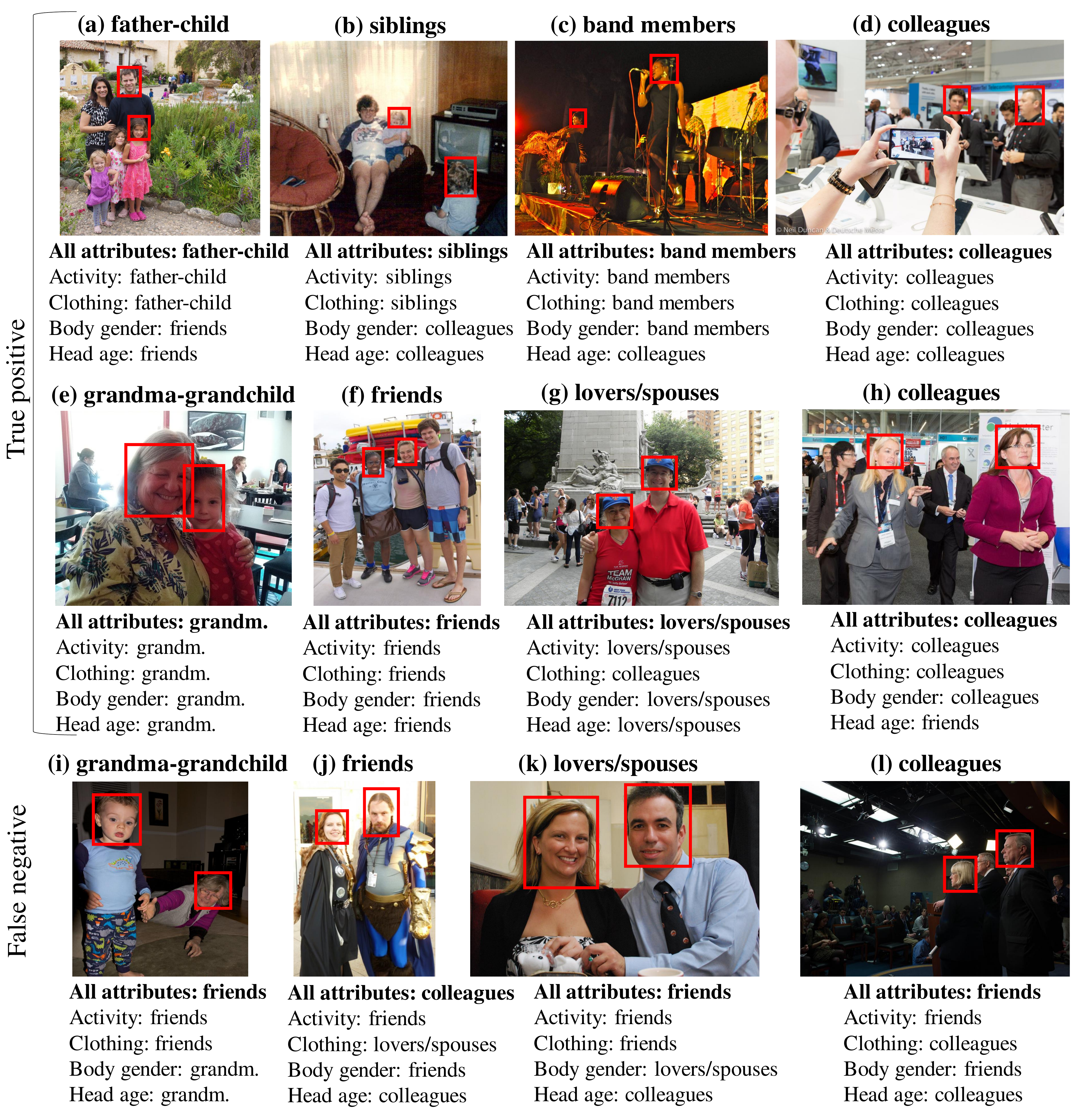}
  \caption{Positive and negative examples of relation recognition. Please note, ``positive'' and ``negative'' here refer to the best results using all-attribute model. Image titles are the groundtruth. Predicted relations using either all-attribution model or top 4 attribute models (ranked in Figure~\ref{normalizedcor}) are listed under images.}\label{baseline-C}
  \label{quality}
\end{figure}

\textbf{Detail attributes}. To figure out detail attributes contributing to social relation recognition, we show some positive and negative examples in Figure~\ref{quality}. Predictions by all-attribute model (our best) and top 4 single-attribute models (ranked in Figure~\ref{normalizedcor}) are listed under images.

Among positive examples, (d)(e)(f) are correctly predicted by all-attribute and 4 single-attribute models. These images contain many strong social cues, \eg in (e), ``large age difference'', ``senior-aged woman'', ``little girl'', ``daily clothing'' and ``intimate hugging'' contribute to the recognition of grandma-grandchild. It is noted that the person pair in (d) are almost in the hazy background but are correctly recognized, validating the ability of our attribute models to handle tough images. Other positive examples in (a)(b)(c)(g)(h) are correctly recognized by all-attribute model but get noisy predictions by single-attribute models, \eg in (g), clothing model makes a prediction of colleagues, probably because this couple wear unified clothes for taking part in an event. Other strong cues like the intimate activity of ``arm on shoulder'', body proximity, age and gender features contribute to the prediction of lovers/spouses when combining all attributes.

Overall, negative examples show more unusual behaviors and confused visual cues than positive ones. For example, the groundtruth of (i) is grandma-grandchild, while the activities of ``grandma crawling and trying to hold baby's hand'' seem very unusual. Although using age or gender finntuned model make correct predictions, activity and clothing are so dominant that the prediction by all-attribute model is wrong. In another example (l), human annotators can recognize the official colleagues in a press conference, but the prediction of our model is failure due to unclear body/head appearances in the image.

\section{Conclusion}
In this paper, we explore a challenging problem of recognizing social relations in daily life photos. While prior work remained partial in terms of the covered social relations, we argue for a social domain based approach in order to investigate relations covering all aspects of social life. We evaluate recognition performance of social relations, social domains as well as generalization performance of recognizing domain across relations.
Our experiments highlight the importance of using semantic attributes, which in turn lends to inspectable models that connect to the social psychology theory.
Beyond the presented work, we belief that our work can pave the way to a more empirical study of social relations that is yet grounded and interpretable in the context of social psychology theories.

\vspace{-0.1cm}
\section*{Acknowledgements}
\vspace{-0.1cm}
This research was supported by the German Research Foundation(DFG CRC 1223).

\newpage
\section*{Supplementary Materials}

\begin{figure*}[hp]
  \centering
  \includegraphics[width=17cm]{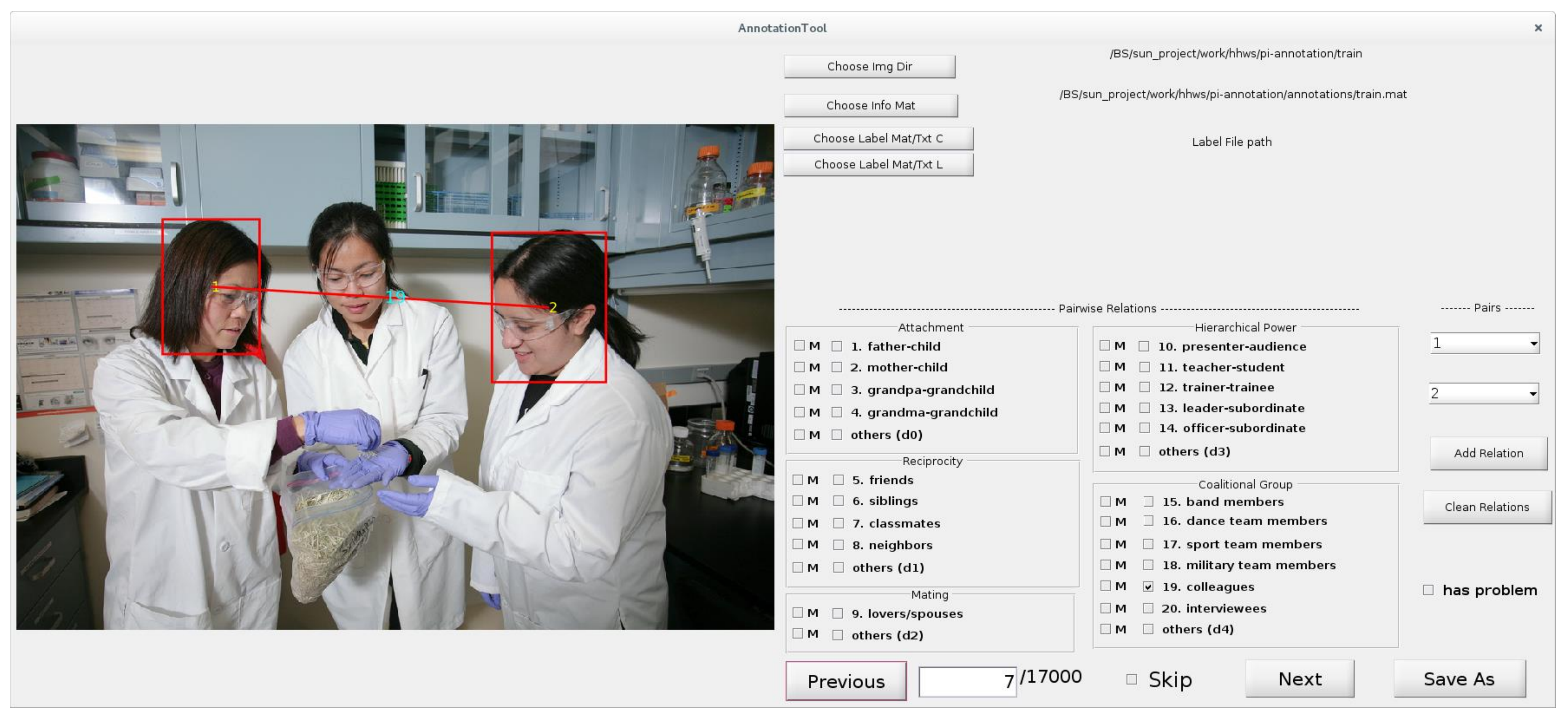}\\
  \caption{The panel of the annotation tool used for the annotation process. Each time the tool pops up a photo with a pair of head bounding boxes. An annotator recognizes the relation based on obvious visual cues, such as age, gender, clothing, activity and so on, then chooses at most 3 relation labels. ``M'' denotes ``maybe'', i.e., not very confident.}\label{annotationTool}
\end{figure*}

\subsection*{Annotation tool} In Figure~\ref{annotationTool}, we present the panel of the annotation tool. Annotators labeled the whole PIPA dataset using this tool, independently. Each time the tool shows a photo with a pair of head bounding boxes\footnote{Head bounding boxes were previously annotated in PIPA for the task of person identification.}. Please note, photos containing only one identity or more than 5 identities\footnote{Identity labels were previously annotated in PIPA for the task of person identification. There are about 3.7\% images containing more than 5 identities, which are mostly same-relation person pairs in groups.} are unused. Therefore, each photo contains 1 $\sim$ 10 person pairs.

On this panel, we have 20 social relation candidates and use ``others'' for annotating some pairs possibly belonging to a domain but not belonging to any relation in this domain. When all annotations were finished, we analysed label statistics and agreements, then filtered out 4 relations (NO. 8, 14, 18, and 20 in this panel) due to insufficient occurrence or insufficient agreement, as explained in Section~\ref{labelList} of the main paper.

\subsection*{Annotation samples}
\begin{figure*}[t]
  \centering
  \includegraphics[width=16cm]{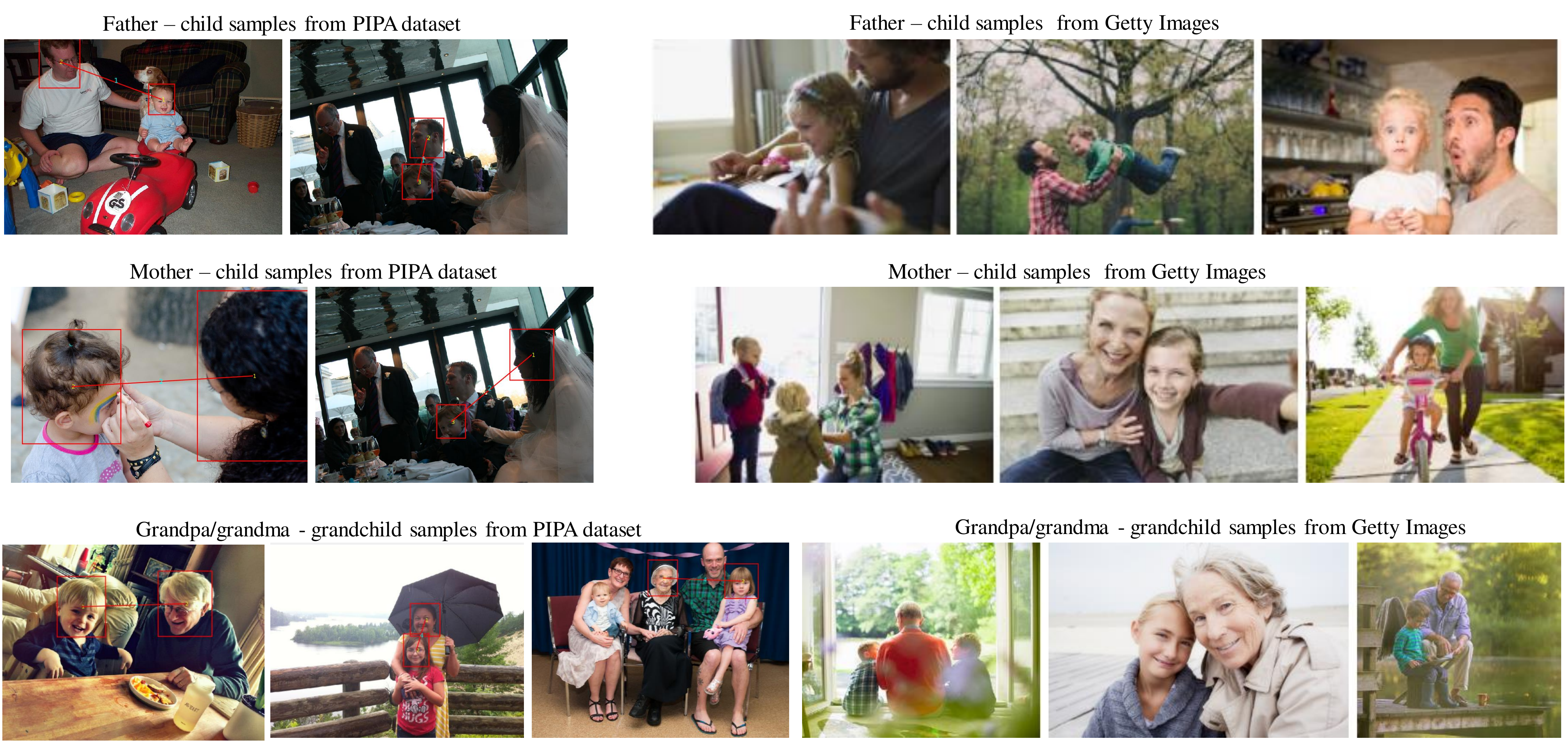}\\
  \caption{Attachment domain: annotation samples from PIPA dataset and searched samples from Getty Images.}\label{attachment}
  \end{figure*}

\begin{figure*}[t]
  \centering
  \includegraphics[width=16cm]{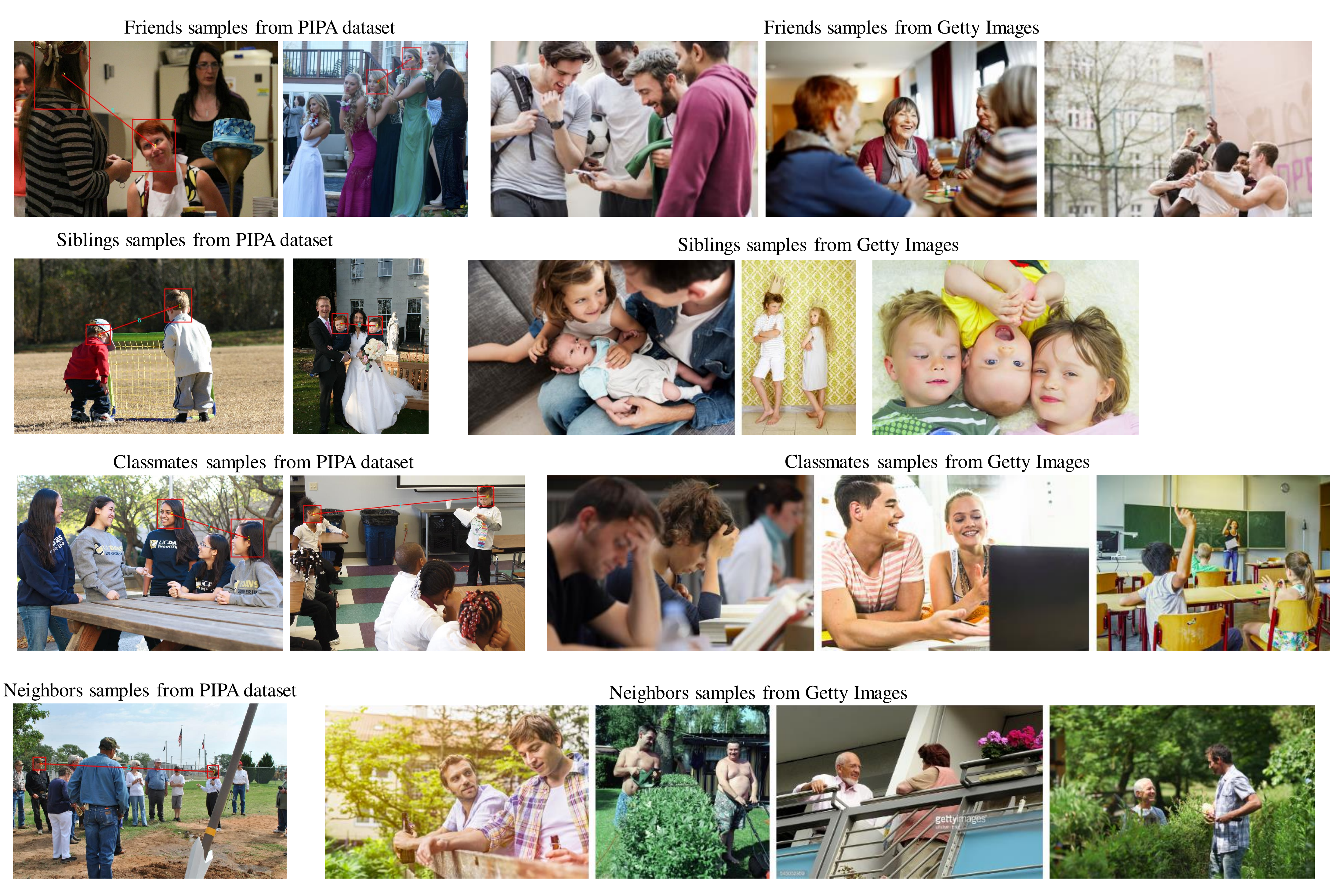}\\
  \caption{Reciprocity domain: annotation samples from PIPA dataset and searched samples from Getty Images.}\label{recip}
\end{figure*}

\begin{figure*}[t]
  \centering
  \includegraphics[width=\linewidth]{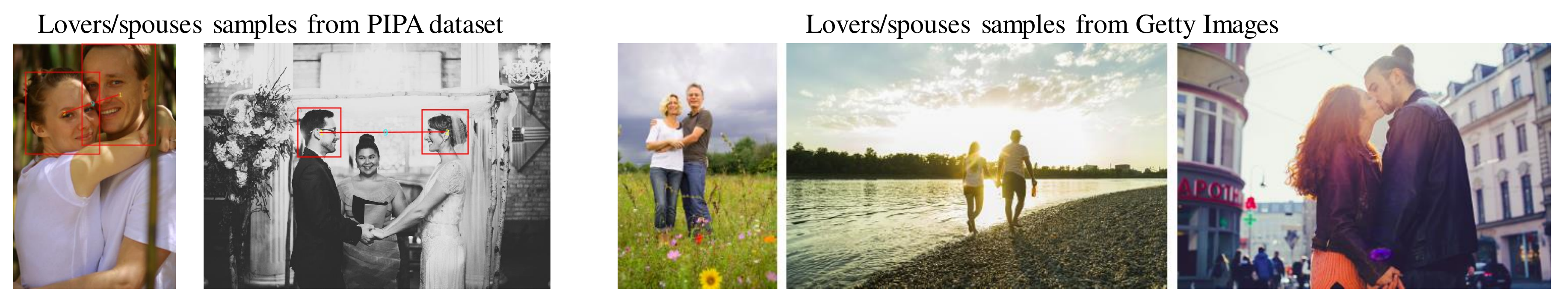}\\
  \caption{Mating domain: annotation samples from PIPA dataset and searched samples from Getty Images.}\label{mating}
\end{figure*}

\begin{figure*}[t]
  \centering
  \includegraphics[width=16cm]{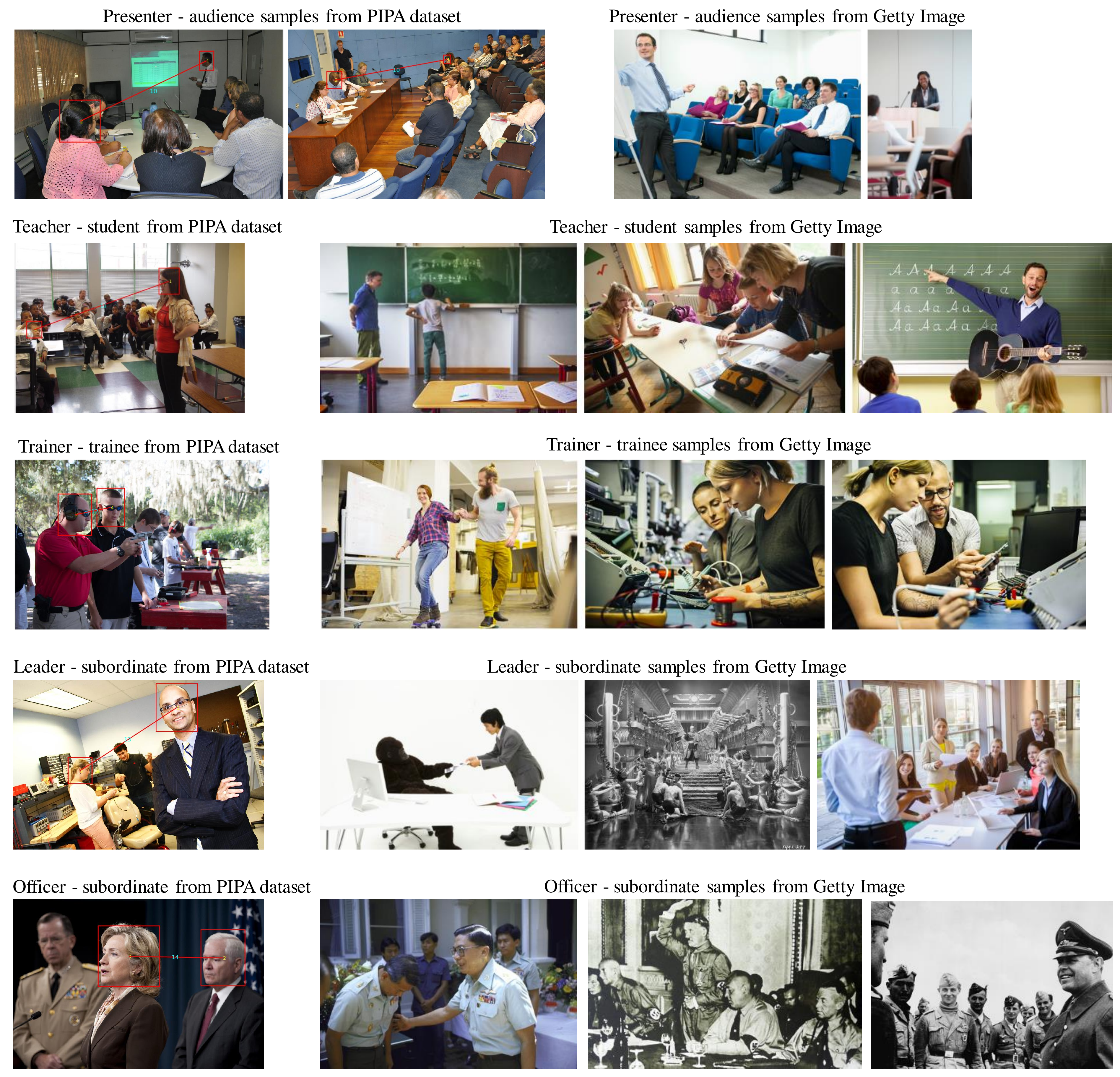}\\
  \caption{Hierarchical power domain: annotation samples from PIPA dataset and searched samples from Getty Images.}\label{hier}
\end{figure*}

\begin{figure*}[t]
  \centering
  \includegraphics[width=16cm]{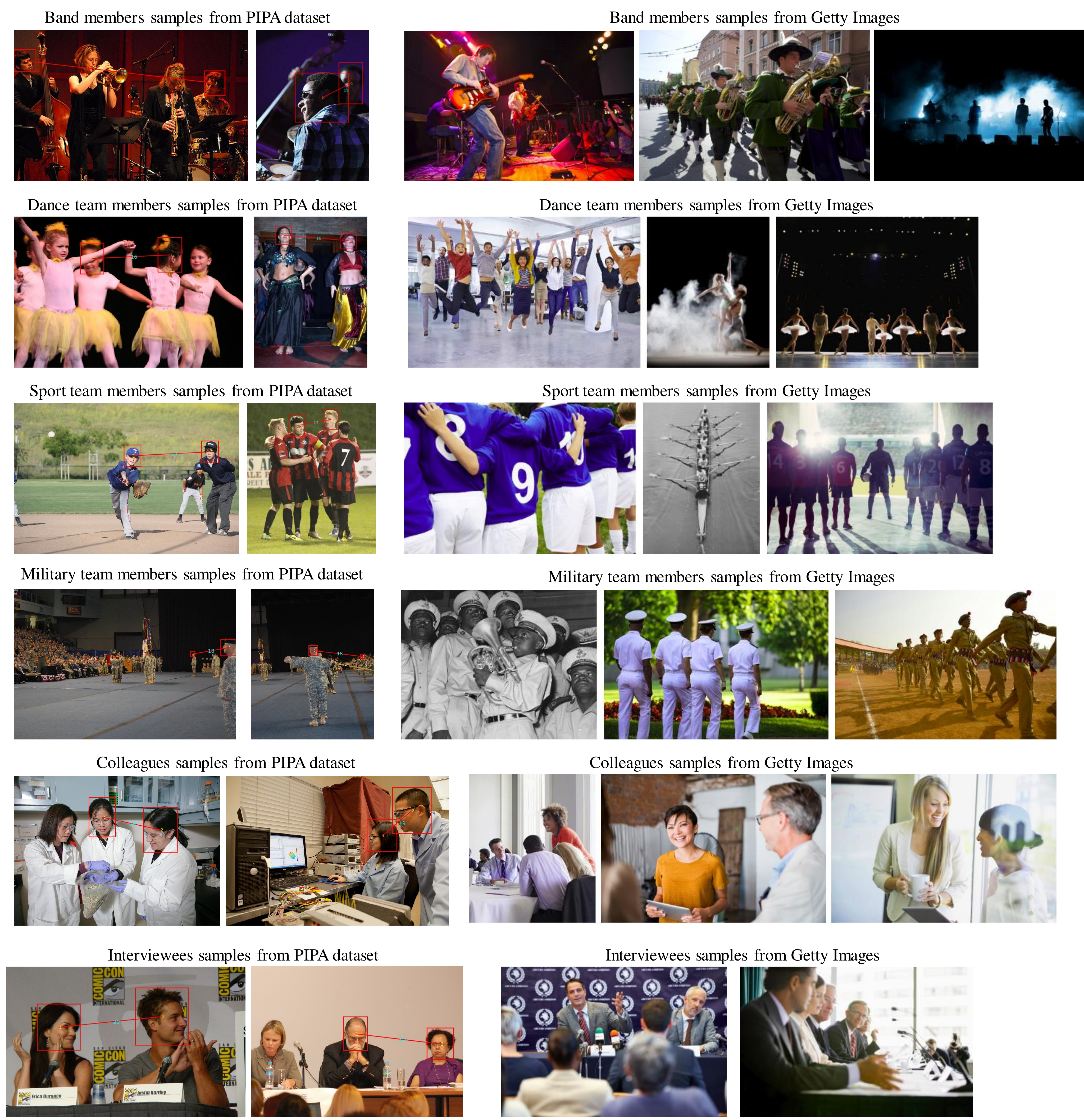}\\
  \caption{Coalitional groups domain: annotation samples from PIPA dataset and searched samples from Getty Images.}\label{coali}
\end{figure*}

Based on the definitions in the social psychology article \cite{Bugental2000}, we gave detailed explanations for 5 social domains. For example, in the Attachment domain, specific visual cues are the age difference between child and parents (or grandparents), body proximity, and children's behavior of seeking protection from adults (see Section~\ref{socialdomaintheory}). We carefully explained these descriptions to 5 annotators, in conjunction with the reference samples shown in Figures~\ref{attachment}-\ref{coali}. 

In Figures~\ref{attachment}-\ref{coali}, we show the reference samples of 20 social relations used for the annotation. PIPA samples are collected from the 10\% PIPA data when choosing a relation label list (see Section~\ref{labelList}). To help annotators to understand social relations, we also borrow some pictures searched in Getty Images, as shown beside PIPA samples. Photos searched on Getty Images are mostly fake and posed for photograph, aiming to reveal the inherent concepts in photos. Getty image samples searched by social relation entries turned out to be very helpful for annotators to better understand the visual appearances of social relations.


\subsection*{Supplementary results}

\begin{figure*}[t]
  \centering
  \includegraphics[width=15cm]{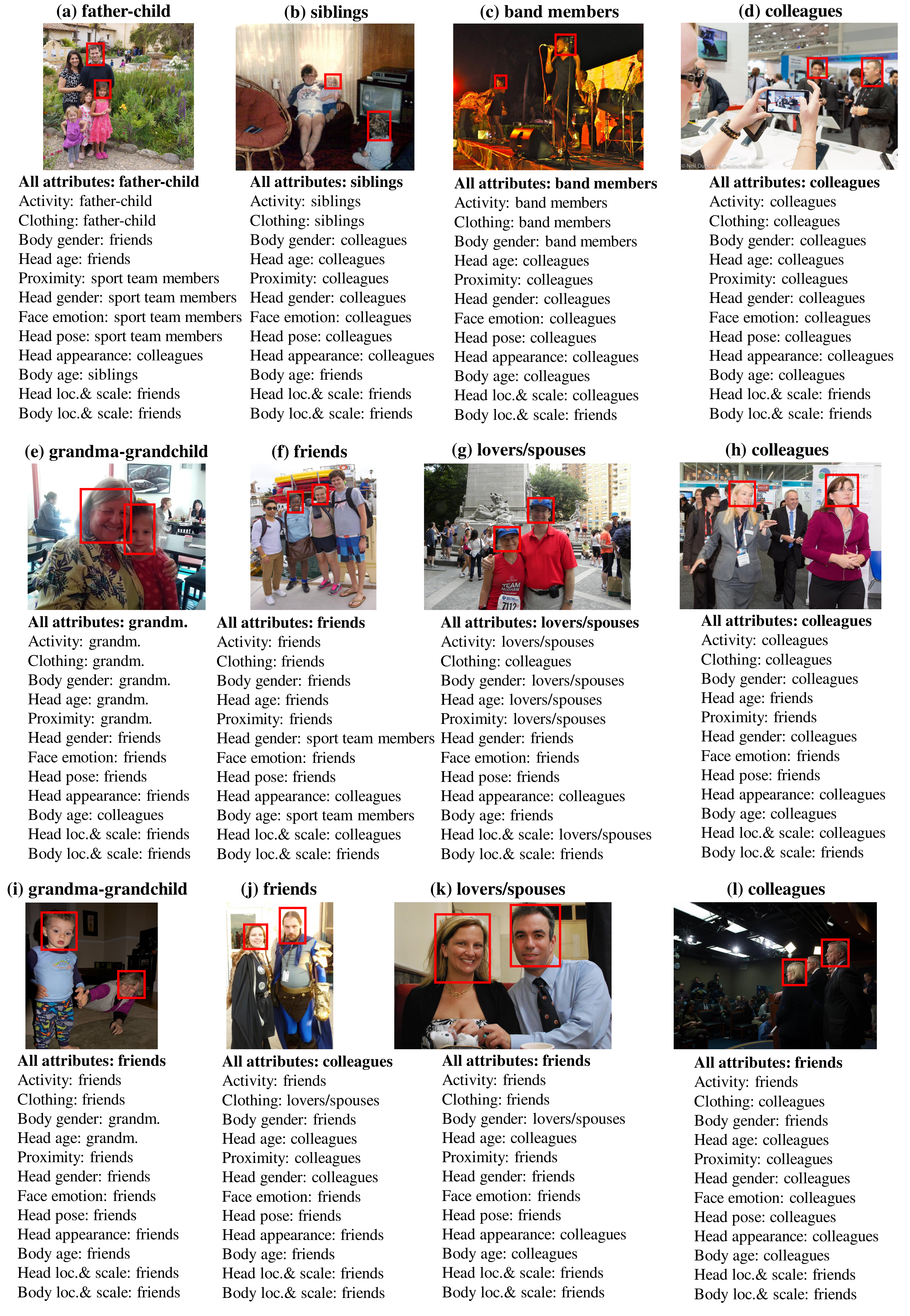}\\
  \caption{Relation labels predicted by the all-attribute model and 12 single-attribute models. Image titles are the ground truth.}\label{results}
\end{figure*}

In our main paper, Figure~\ref{quality} showed the all-attribute result (our best) and top 4 single attribute results (contribution ranks were shown in the Figure~\ref{normalizedcor}). In this supplementary document, Figure~\ref{results} supplements the relation labels predicted by other 8 single-attribute models, i.e., from ``proximity'' to ``body loc.\& scale'', under images. Here, these images are in the same order with those in the Figure~\ref{quality} of the main paper, and the image titles are the ground truth. Positive samples are Figure~\ref{results}(a)-(h) which are correctly predicted by our all-attribute model while get noisy predictions by single-attribute models. For example in Figure~\ref{results}(a), father-child relation was wrongly predicted by 10 single-attribute models except ``activity'' and ``clothing''. This is consistent with the contribution ranks shown in the Figure 6 of the main paper that the attributes of ``activity'' and ``clothing'' dominate the recognition. Negative samples are shown in Figure~\ref{results}(i)-(l).

To plot the Figure~\ref{normalizedcor} in the main paper, we gave an example of computing X, Y coordinates of ``body age'' by defining $X = \text{\it acc}$(bodyAge, domain)$/\text{\it acc}$(all, domain) and $Y = \text{\it acc}$(bodyAge, relation)/$\text{\it acc}$(all, relation). In this document, we supplement all accuracies $\text{\it acc}$ used for plotting, in Table~\ref{recog-results}. Taking the previous example ``body age'', its coordinates were computed as $X = 57.4\% / 67.8\% = 0.847$, $Y = 31.0\% / 57.2\% = 0.542$.

\renewcommand\arraystretch{1.2}
 \begin{table}[h]
 \scriptsize
   \begin{center}
     \label{Atts}
     \begin{tabular}{lccc}
       \toprule
{\sc Attribute} & {\sc Relation recognition} & {\sc Domain recognition} \\ \hline
Head age          &42.8\% & 56.8\% \\
Head gender       &38.0\% & 53.8\% \\
Head loc.\& scale &30.8\% & 45.0\% \\
Head appearance   &31.5\% & 48.4\% \\
Head pose         &34.7\% & 52.3\% \\
Face emotion      &37.7\% & 55.3\% \\
Body age    &31.0\% & 57.4\% \\
Body gender &46.6\% & 58.0\% \\
Body loc.\& scale &27.7\% & 44.2\% \\
Clothing   &51.4\% & 60.3\% \\
Proximity  &39.6\% & 55.4\% \\
Activity   &52.4\% & 64.5\% \\ \hline
All Attributes  &57.2\% & 67.8\% \\
        \bottomrule
     \end{tabular}
     \vspace{0.1cm}
       \caption{Accuracies of recognizing relations and domains using single-attribute models and all-attribute model. These numbers were used to compute the coordinates of attribute dots presented in Figure~\ref{normalizedcor}.}
       \label{recog-results}
   \end{center}
   \vspace{-0.8cm}
 \end{table}

\end{document}